\documentclass[conference]{IEEEtran}
\IEEEoverridecommandlockouts
\usepackage{cite}
\usepackage{amsmath,amssymb,amsfonts}
\usepackage{algorithmic}
\usepackage{graphicx}
\usepackage{textcomp}
\usepackage{xcolor}
\usepackage{booktabs}

\usepackage{listings}
\lstdefinelanguage{yaml}{
  keywords={true,false,null,y,n},
  sensitive=false,
  comment=[l]{\#},
  morestring=[b]',
  morestring=[b]"
}
\def\BibTeX{{\rm B\kern-.05em{\sc i\kern-.025em b}\kern-.08em
    T\kern-.1667em\lower.7ex\hbox{E}\kern-.125emX}}
\begin{document}

\title{Self-Adaptive Anomaly Detection with Reinforcement Learning and Human Feedback in Connected Vehicles\\
}

\author{
\IEEEauthorblockN{Matthias Weiß, Athreya Hosahalli Prakash, Maurice Artelt, Falk Dettinger, Nasser Jazdi and Michael Weyrich}
\IEEEauthorblockA{
\textit{Institute of Industrial Automation and Software Engineering (IAS)} \\
\textit{University of Stuttgart} \\
Pfaffenwaldring 47, 70550 Stuttgart, Germany \\
E-Mail: \{vorname.nachname\}@ias.uni-stuttgart.de}}

\maketitle

\begin{abstract}
Connected vehicles are autonomous cyber-physical systems whose behavior must be continuously monitored during operation to detect deviations from normal operation before they propagate into failures. Such evaluation is challenging because the systems themselves evolve: over-the-air updates, configuration changes, and shifting workloads alter the definition of normal behavior, causing static diagnostic methods to degrade silently over time. Existing approaches typically address either automated model adaptation or operator integration in isolation, rather than as a single coordinated supervisory loop.

This paper presents an online anomaly detection framework for autonomous CPS that integrates three coordinated mechanisms. A factorized deep Q-network with self-attention selects the most suitable detector from a candidate pool for each monitored service, exploiting inter-service dependencies in the microservice topology. An ensemble of three statistical drift detectors monitors the input distribution and raises an alarm only when all three concur, prioritizing precision over recall. A human-in-the-loop retraining mechanism, built around a pending transition buffer and a 60/40 prioritized replay strategy, allows the operator to incorporate expert knowledge while preserving the system's learned response to prior data distributions.

The framework is evaluated on a connected-vehicle testbed running an automated valet parking application across seven backend microservices. The attention-augmented agent achieves an F1 score of 0.69, compared to at most 0.11 for any single detector applied uniformly. Following a real software update that induces measurable concept drift, F1 drops to 0.52; after operator-triggered retraining, performance recovers to 0.65 on the new distribution while remaining at 0.69 on the prior one, demonstrating sustained adaptation without catastrophic forgetting.
\end{abstract}


\section{Introduction}
Connected vehicle (CV) functions such as over-the-air updates, remote diagnostics, real-time traffic optimization, or cooperative maps, are increasingly deployed in production vehicles, transforming the vehicle from a standalone product into one node in a distributed cyber-physical system spanning cloud services, edge nodes, and other vehicles \cite{guo2025automated, Stuempfle2025SDV, dettinger2025directives}. These systems operate with growing autonomy: they coordinate among themselves, allocate computational resources, and respond to environmental conditions with limited operator involvement \cite{praveen2025autonomous, dettinger2024future}. As autonomy grows, so does the need to continuously evaluate operational behavior, in order to detect deviations before they propagate into service-affecting or safety-relevant failures \cite{weiss2024simulating, lai2024kansai}.

Behavior evaluation in such systems is challenging because the systems themselves evolve. Frequent over-the-air updates, configuration changes, and shifting workloads alter the statistical properties of monitored signals, creating concept drift that silently degrades static diagnostic methods over time \cite{weiss2023continuous}. Compounding this, the volume and heterogeneity of signals across cloud, edge, and vehicle domains exceed what manual operator inspection can reasonably cover \cite{lim2021state}. Diagnostic methods must therefore scale with data dimensionality and adapt continuously, without losing the system's prior learned behavior.

Existing solutions tend to fall into one of two camps. Automated adaptation approaches such as model-selection strategies \cite{zhang2022time, weiss2024ad} respond to changing data distributions but operate as black boxes, leaving operators with limited ability to inject domain knowledge or override mistaken inferences. Operator-in-the-loop approaches, conversely, retain human authority but lack mechanisms to detect when adaptation is needed or to preserve prior knowledge during retraining. For autonomous CPS where both continuous adaptation and operator oversight are required, neither category suffices on its own.

This paper presents an self-adaptive anomaly detection framework that closes this gap through three coordinated mechanisms:
\begin{itemize}
    \item A \textbf{factorized deep Q-network with self-attention} that selects an appropriate detector from a candidate pool for each monitored service, exploiting inter-service dependencies in the microservice topology.
    \item An \textbf{ensemble of three statistical drift detectors} with a conjunctive alarm rule that prioritizes precision and signals to the operator when retraining is warranted.
    \item A \textbf{human-in-the-loop retraining mechanism} using a pending transition buffer and a prioritized replay strategy that integrates expert feedback while preventing catastrophic forgetting.
\end{itemize}
The framework is integrated into the SDVDiag diagnostic platform \cite{weiss2025sdvdiag} and evaluated on a connected-vehicle testbed running an automated valet parking application across seven backend microservices.

The remainder of this paper is structured as follows: Section~\ref{sec:related_work} reviews related work on anomaly detection, concept drift, and human-in-the-loop adaptation in connected vehicle environments. Section~\ref{sec:concept} details the proposed framework. Section~\ref{sec:evaluation} presents the evaluation on the testbed. Finally, Section~\ref{sec:conclusion} concludes and outlines future work.

\section{Related Work}\label{sec:related_work}

Adapting anomaly detection to non-stationary data streams has produced several non-RL approaches~\cite{bayram2022concept, weiss2025review}. The Adaptive Windowing algorithm (ADWIN)~\cite{das2024aidps} maintains a variable-length window of recent observations and uses statistical tests to detect distribution changes, automatically discarding outdated data. The Split Active Learning Anomaly Detection (SALAD) method~\cite{lee2021salad} combines autoencoders with active learning and fading factors to address gradual and abrupt drift while minimizing labeling costs. LSTM-based streaming methods~\cite{saurav2018online} leverage temporal dependencies for incremental model updates. These approaches require careful threshold tuning, depend on periodic human annotation, or lack explicit drift-handling mechanisms, and none provide a structured channel for operator knowledge during operation.

Reinforcement learning has been applied to anomaly and intrusion detection in several formulations that treat the agent as a classifier over individual observations~\cite{sedar2022reinforcement, wang2022deep, he2024reinforcement, yoon2024intrusion}. Closer to our setting, Zhang et al.~\cite{zhang2022time} proposed an RL-based model-selection approach for time-series anomaly detection, in which the agent learns to choose a detector from a pool based on extracted features. This formulation is the closest prior work to ours; what is missing is online operation under concept drift, an explicit drift-detection mechanism, and operator integration into the training loop. The factorized DQN architecture of Zhou et al.~\cite{zhou2019factorized}, originally developed for general multi-discrete action spaces, provides the structural foundation we adapt to per-service detector selection in microservice topologies.

Statistical drift detection has matured into a substantial subfield of streaming machine learning, with a recent survey by Hinder et al.~\cite{hinder2024drift} cataloging the unsupervised setting most relevant to anomaly-detection pipelines. Sequential change-point methods such as the Page--Hinkley test capture mean shifts, two-sample tests such as Kolmogorov--Smirnov capture changes in marginal distribution shape, and distance-based multivariate methods such as Mahalanobis-distance outlier rates capture shifts in joint dependency structure~\cite{coil2025distance}. The recently proposed ADA-ADF framework~\cite{li2025unsupervised} integrates KS-based drift detection with reconstruction-error monitoring inside a streaming anomaly-detection pipeline and adapts via replay-ratio adjustment---an architecturally adjacent approach that nonetheless addresses neither RL-based detector selection nor expert feedback. In parallel, reinforcement learning from human feedback has emerged as a structured mechanism for incorporating expert knowledge into training~\cite{kaufmann2025rlhf}, with recent work demonstrating human-in-the-loop frameworks for time-series anomaly detection~\cite{deng2024hilad}; these efforts, however, remain disconnected from drift detection and from the autonomous CPS supervisory context.

To the best of the authors' knowledge, no prior work integrates RL-based detector selection, statistical drift detection, and operator-driven retraining with forgetting prevention into a single supervisory loop for connected vehicle anomaly detection.

\begin{figure*}[!tb]
    \centering
    \includegraphics[width=\linewidth]{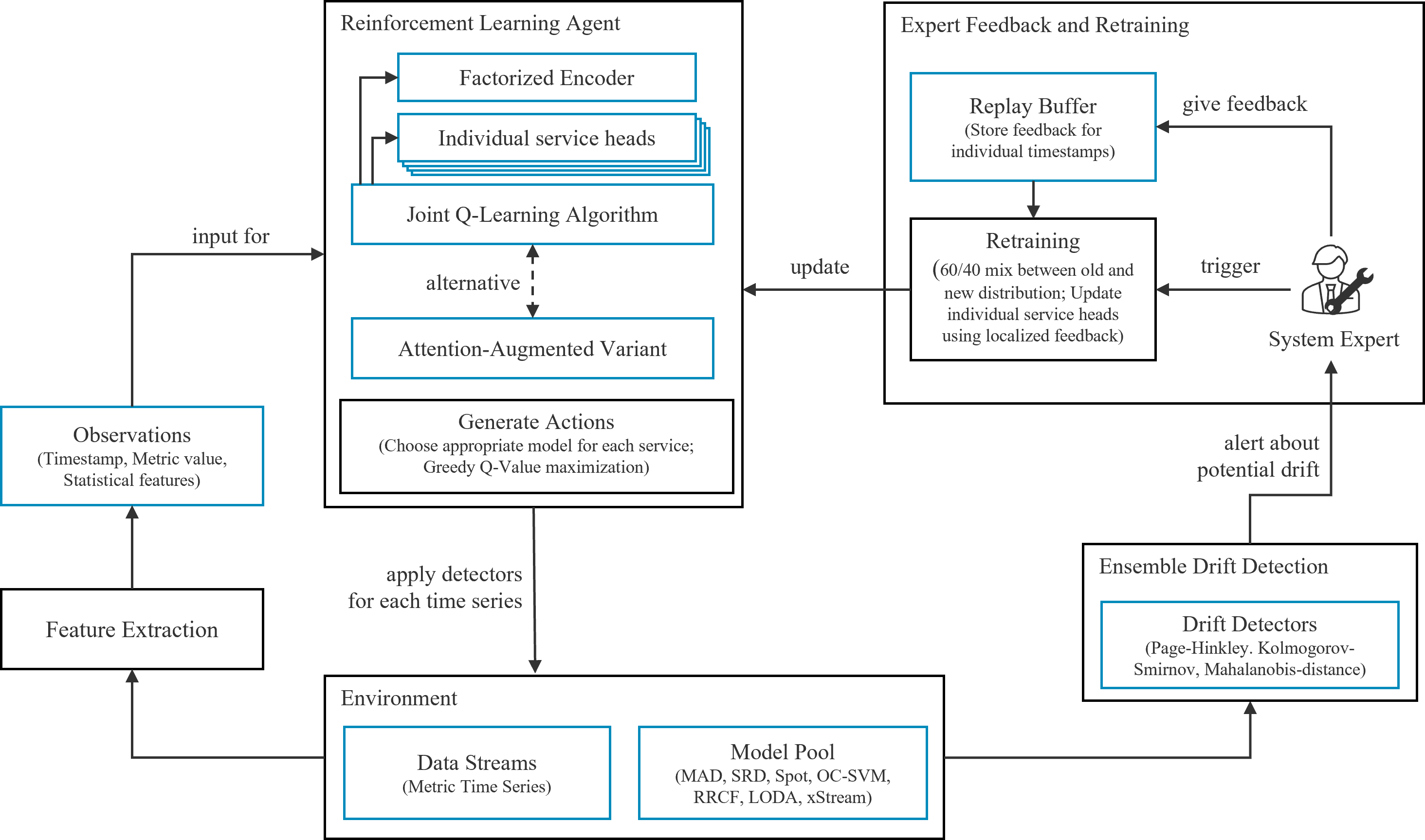}
    \caption{Architecture of the proposed operator-supervised anomaly detection framework. Statistical features extracted from per-service metric streams are passed to a factorized RL agent (MLP or attention-augmented variant), which selects an appropriate detector from the model pool for each service. In parallel, an ensemble of three drift detectors monitors the input distribution and alerts the system expert when all three concur. The expert validates flagged anomalies and triggers retraining via a replay buffer that mixes recent and prior transitions to prevent catastrophic forgetting.}
    \label{fig:core_arch}
\end{figure*}

\section{Concept for Online Anomaly Detection with Expert Feedback}\label{sec:concept}
The proposed approach establishes a reinforcement learning based model selection loop that integrates automated concept drift detection and re-training using expert feedback. At the center of the concept is a deep reinforcement learning agent trained to select the optimal anomaly detector from a pool of candidate models for each monitored time series. By learning which detector performs best under varying data characteristics, the agent effectively mimics the decision-making process of a system expert while enabling continuous adaptation to changing operational conditions. The resulting framework is integrated into the SDVDiag platform \cite{weiss2025sdvdiag}, a diagnostic infrastructure designed for connected vehicle environments. SDVDiag provides capabilities for collecting operational metrics from the vehicle testbed, applying anomaly detection in real time, and utilizing detected anomalies in subsequent analysis steps such as root cause identification.

Figure~\ref{fig:core_arch} presents an overview of the proposed approach. The architecture comprises four principal components: a layer for metric collection and feature extraction within the environment, the reinforcement learning agent, which is trained to select the most appropriate anomaly detector based on the generated statistical features, an ensemble of statistical concept drift detection modules, and finally an expert feedback interface, which provides the ability to fine tune the detector performance and trigger a retraining if concept drift is detected. The following subsections provide details for each component.

\subsection{Data Integration and Feature Extraction}\label{subsec:data_integration}

Operational metrics are collected from each backend microservice using an OpenTelemetry-based monitoring stack deployed within the connected vehicle infrastructure. For the evaluation environment described in Section~\ref{sec:evaluation}, two metrics are collected per service: CPU utilization and memory (RAM) usage, both sampled at one-second resolution. These metrics are normalized to the range $[0, 1]$ prior to processing.

The raw metric stream for each service is processed through a sliding window of fixed length $W$, which advances with each new observation. For each window, a set of nine statistical descriptors is computed: mean, standard deviation, minimum, maximum, skewness, kurtosis, slope (estimated by linear regression over the window), the first-order difference between consecutive windows (capturing short-term change), and the interquartile range. Together, these features form a compact yet expressive representation of the current data condition, enabling the agent to distinguish between stable, trending, or high-variance regimes and adapt its detector selection accordingly.

This feature vector is then passed to the RL agent as its state observation. The pool of candidate anomaly detectors available for selection is designed to contain methods suitable for various different data regimes. It includes: MAD (median absolute deviation), SRD (spectral residual), Spot (peaks over threshold), OC-SVM (one-class SVM), RRCF (robust random cut forest), LODA (lightweight online detector of anomalies) and xStream.

\subsection{Reinforcement Learning Loop}\label{subsec:rl_loop}
After receiving the state observation, the agent is tasked with selecting the optimal anomaly detector for each monitored time series. 

\subsubsection{Problem Formulation}

The detector selection problem is formulated as a multi-discrete Markov Decision Process. At each timestep, the agent observes the current state $s$ (the concatenated feature vectors for all monitored services), selects an action $\mathbf{a} = (a_1, \ldots, a_N)$ consisting of a detector choice $a_i$ for each of $N$ services, and receives a reward $r$ reflecting the quality of the selected detectors' outputs

\subsubsection{Factorized Network Architecture}

A standard DQN architecture would require an output layer of size $|D|^N$, where $|D|$ is the number of candidate detectors, making it computationally intractable as the number of services or detectors grows. Furthermore, when the service topology changes (e.g., a new service is added or removed), the input and output dimensions change, requiring complete retraining.

To address these challenges, a factorized architecture inspired by Zhou et al.~\cite{zhou2019factorized} is employed. It consists of two components: a shared encoder and independent per-service heads. The shared encoder processes the full input state through a stack of fully connected layers with ReLU activations, extracting a high-level feature representation relevant across all services. This shared representation is then passed to $N$ independent linear heads, one per service, each mapping the shared features to Q-values over the $|D|$ possible detector choices for that service.

When a service is added or removed, only the corresponding head needs to be added or retrained; the shared encoder and all other heads remain intact. This provides a significant practical advantage in evolving microservice architectures.

\subsubsection{Joint Q-value Learning}

Training is performed using a joint Q-value decomposition strategy. The total Q-value is decomposed as:
\begin{equation}
    Q_{\text{joint}}(s, \mathbf{a}) = \sum_{i=1}^{N} Q_i(s, a_i)
\end{equation}
where $Q_i(s, a_i)$ is the Q-value produced by head $i$ for selecting detector $a_i$. During training, the target value follows the same decomposition:
\begin{equation}
    y_{\text{joint}} = r + \gamma \sum_{i=1}^{N} \max_{a'_i} Q_i(s', a'_i)
\end{equation}
The network is updated by minimizing the mean squared error:
\begin{equation}
    \mathcal{L} = \left( y_{\text{joint}} - Q_{\text{joint}}(s, \mathbf{a}) \right)^2
\end{equation}
This loss updates the shared encoder jointly, while each service head is additionally updated using its individual Q-values, allowing each head to learn its own specialized contribution.

\subsubsection{Attention-Augmented Variant}

The standard factorized architecture (F-DQN) treats each service's features independently before aggregating at the shared encoder. However, in a microservice architecture, services are not independent: a CPU spike in the central coordination service is likely to cause cascading effects in dependent services, as visible in the service topology described in Section~\ref{sec:evaluation}. Exploiting these inter-service dependencies enables more contextually informed detector selection.

The attention-augmented variant (F-DQN-Attn) addresses this through a hierarchical processing pipeline. Each service's 9-dimensional feature vector is first encoded independently through a small per-metric encoder, producing a per-service embedding. These $N$ service embeddings are then passed through a multi-head self-attention layer, allowing each service to attend to the embeddings of all other services before producing its final representation. This attended representation is passed to the shared encoder and subsequently to the per-service heads. In contrast to the MLP variant, which flattens the full observation into a single vector with no explicit structure, the attention variant explicitly models the relational structure of the service topology. This is particularly beneficial for detecting correlated failure patterns where an anomaly in one service provides predictive information about the health of connected services.

\subsubsection{Reward Function and Training Initialization}

The agent is initialized through supervised training on a labeled dataset generated by injecting known anomaly patterns --- spikes, gradual drifts, and service degradation scenarios --- into the system. Rewards are assigned based on each detector's output relative to ground truth, using an asymmetric weighting that penalizes false negatives more heavily than false positives (Explicit reward values were determined empirically as TP: +40; FP: -15; FN: -20; TN: +0.1). This reflects the operational reality that missing a genuine anomaly is costlier than issuing a spurious alarm in a system where anomalies are rare. The framework also supports unsupervised operation when labels are unavailable, using Excess Mass and Mass Volume metrics~\cite{goix2016evaluate} as proxy rewards.

An experience replay buffer is employed to break temporal correlations during training, storing transitions $(s, \mathbf{a}, r, s')$ and sampling uniformly during each update step to improve gradient stability and sample efficiency.

Table~\ref{tab:architecture} summarizes the F-DQN-Attn network architecture, and Table~\ref{tab:hyperparams} lists the training hyperparameters used throughout the evaluation. The replay buffer is intentionally sized to retain transitions across multiple data-distribution regimes, supporting the prioritized replay strategy described in Section~\ref{sec:concept}-D-2.

\begin{table}[t]
\caption{F-DQN-Attn architecture. $B$: batch size, $N$: number of services, $M$: number of metrics per service, $K$: number of statistical features per metric.}
\label{tab:architecture}
\centering
\begin{tabular}{lll}
\toprule
\textbf{Layer} & \textbf{Configuration} & \textbf{Output shape} \\
\midrule
Input             & ---                                & $(B, N, M, K)$ \\
Metric encoder    & embed dim 128                      & $(B, N, M, 128)$ \\
Service encoder   & embed dim 128                      & $(B, N, 128)$ \\
Multi-head attn.  & 8 heads, dropout 0.1               & $(B, N, 128)$ \\
Service heads     & 2-layer MLP, $128{\to}128{\to}|D|$ & $(B, |D|) \times N$ \\
\bottomrule
\end{tabular}
\end{table}

\begin{table}[t]
\caption{Training hyperparameters.}
\label{tab:hyperparams}
\centering
\begin{tabular}{ll}
\toprule
\textbf{Parameter} & \textbf{Value} \\
\midrule
Learning rate              & $1 \times 10^{-4}$ \\
LR schedule                & StepLR (20{,}000 steps, $\gamma = 0.5$) \\
Discount factor $\gamma$   & 0.99 \\
Batch size                 & 2{,}048 \\
Replay buffer size         & $2 \times 10^{6}$ \\
Target network update      & every 500 steps \\
$\varepsilon$-schedule     & 1.0 $\to$ 0.1, decay 0.9999 \\
Statistical feature window $W$ & 64 \\
Training frequency         & 1 update / 4 timesteps \\
\bottomrule
\end{tabular}
\end{table}

\subsection{Drift Detection Modules}\label{subsec:drift_detection}
In dynamic connected vehicle environments, the input distribution observed by the RL agent is rarely stationary: software releases, load shifts, and infrastructure reconfigurations can silently alter feature statistics and degrade policy quality without any explicit signal of failure. Since no single statistical test reliably captures every form of distributional shift, the framework employs an ensemble of three complementary detectors, each probing a different geometric property of the data, and raises a drift alarm only when all three concur. This conjunctive rule reflects a deliberate preference for precision over recall, since false alarms in the continuous training loop can trigger unnecessary buffer invalidations, whereas a brief detection delay is tolerable.

The \textbf{Page--Hinkley test} is applied per feature against baseline statistics estimated on an earlier temporal window. It accumulates deviations from the baseline mean, discounted by a tolerance term, and flags features whose cumulative statistic exceeds a threshold proportional to the reference standard deviation. This sequential change-point method is particularly effective for sustained mean shifts and gradual trends.

The \textbf{two-sample Kolmogorov--Smirnov test} is run per feature between two consecutive windows. To avoid the well-known p-value collapse at large sample sizes, the implementation thresholds directly on the D-statistic ($D > 0.15$). It complements Page--Hinkley by detecting any change in the marginal distribution---variance, skewness, or modality---rather than only the mean. For both per-feature tests, a system-level alarm requires that at least $40\%$ of features individually trigger, preventing a handful of noisy dimensions from dominating the verdict.

The third detector, a \textbf{Mahalanobis-distance outlier-rate test}, operates in the joint feature space. The reference window provides a mean vector and a ridge-stabilized covariance matrix; points beyond the 99th percentile of the corresponding chi-squared distribution are flagged as outliers. Drift is reported when the query window's outlier rate both doubles relative to the baseline and exceeds it by an absolute margin of $0.5$, ensuring the alarm reflects a genuine structural shift rather than heavy-tailed reference data.

Together, the three methods form a layered diagnostic: Page--Hinkley guards against mean drift, Kolmogorov--Smirnov against marginal shape change, and Mahalanobis against shifts in multivariate dependency.

\subsection{Expert Feedback and Retraining}\label{subsec:human_feedback}

In operational settings, human experts act as critical collaborators alongside the automated detection pipeline. Through the feedback interface, operators monitor real-time streaming time series and their corresponding anomaly scores. When the system flags a potential anomaly, the expert can validate or correct it. A deliberately pessimistic default is applied: any flagged anomaly is treated as a false positive during subsequent retraining unless the expert explicitly validates it. This conservative stance combats alert fatigue by ensuring that model updates are driven exclusively by verified, high-confidence signals.

To handle the asynchronous nature of human feedback, a pending transition buffer is maintained. Incomplete transitions (state and action, without reward) are stored indexed by a timestamp-based key. When the expert submits feedback, the corresponding transition is retrieved, the reward is attached, and the complete transition $(s, \mathbf{a}, r, s')$ is migrated to the main replay buffer for training.

Inserting a completed transition does not automatically trigger retraining. Control remains with the human operator, who initiates retraining explicitly. To guide this decision, the interface surfaces unanimous drift alarms from the ensemble drift detection module, alerting the expert when the underlying data distribution has materially changed and retraining is likely to be beneficial.

\subsubsection{Selective Head Update for Localized Feedback}

A practical challenge in the factorized architecture is that expert feedback is typically localized: an operator investigating a specific anomaly will annotate one or a few services, leaving others unannotated. Updating only the heads with explicit feedback while freezing others would deprive the shared encoder of a complete gradient signal.

To address this, a selective update strategy is employed. The head corresponding to the annotated service receives the explicit reward derived from the expert's feedback. Remaining heads, lacking explicit annotation, are assigned pseudo-rewards assuming that the agent's original predictions were correct (i.e., true positives or true negatives as appropriate). This allows a complete joint loss to be computed for updating the shared encoder while keeping the annotation burden on the expert realistic.

\subsubsection{Prioritized Replay for Catastrophic Forgetting Prevention}

During retraining, the replay buffer is sampled using a prioritization strategy: A portion of each training batch is drawn from transitions collected under the new data distribution, whereas the remaining samples are transitions collected under the prior distribution. This deliberate imbalance ensures that the model adapts to the new distribution at a faster rate while continuously rehearsing prior knowledge, preventing catastrophic forgetting. For the purpose of this paper, a split of 60/40 (new data/old data) was derived empirically to provide the best compromise between adaptation speed and stability.  The effectiveness of this strategy is demonstrated in Section~\ref{subsec:human_feedback_eval}.

\section{Evaluation}\label{sec:evaluation}

\begin{figure}[!b]
    \centering
    \includegraphics[width=0.9\linewidth]{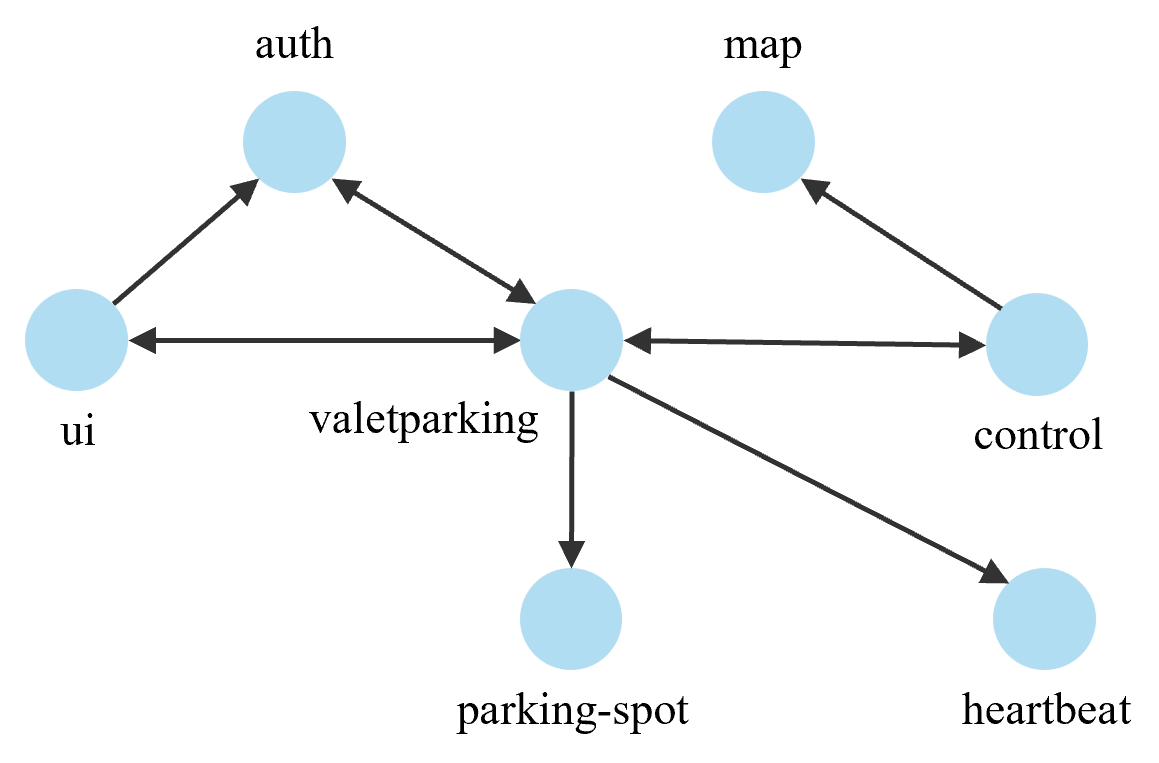}
    \caption{Architecture of the AVP backend microservices. The \texttt{valetparking} service acts as the central coordinator, communicating with \texttt{auth} and \texttt{ui} for access control and user interaction, \texttt{control} and \texttt{map} for vehicle command routing and spatial data, and \texttt{parking-spot} and \texttt{heartbeat} for slot management and vehicle health monitoring.}
    \label{fig:services}
\end{figure}

The proposed framework has been implemented and fully integrated into the SDVDiag platform, deployed within the \textbf{intelligent fleet} operated at the University of Stuttgart. The testbed includes four unmanned ground vehicles (UGVs) equipped with comprehensive sensor suites for autonomous navigation and 5G cellular modules for wireless connectivity. A central Kubernetes server backend consisting of five computing nodes hosts the connected vehicle functions.

The function under evaluation is an \textbf{Automated Valet Parking (AVP) application}, where the UGVs are tasked with navigating to designated parking spots and drop/retrieve locations based on user requests. In this scenario, the allocation and management of available parking spaces is handled by cloud backend microservices. Each UGV navigates autonomously between designated parking spots and passenger zones, requiring continuous synchronization between vehicle and cloud components.

The backend architecture is depicted in Figure~\ref{fig:services}. It consists of seven microservices: \texttt{valetparking} (central coordinator), \texttt{auth} (authentication), \texttt{ui} (user interface), \texttt{control} (vehicle command routing), \texttt{map} (spatial data provider), \texttt{parking-spot} (parking slot management), and \texttt{heartbeat} (vehicle health monitor). The topology follows a hub-and-spoke pattern centered on the \texttt{valetparking} service, with bidirectional communication to \texttt{auth}, \texttt{ui}, and \texttt{control}, and outgoing connections to \texttt{map}, \texttt{parking-spot}, and \texttt{heartbeat}. From each service, CPU utilization and RAM usage are collected at one-second resolution and a sliding window size of 64 events via the OpenTelemetry monitoring stack, yielding 14 monitored time series in total (2 metrics $\times$ 7 services).

The following subsections present the evaluation results. Section~\ref{subsec:raw_performance} compares the performance of the pre-trained RL agent with individual baseline detectors to verify that the selections of the agent mirror expert decisions. Section \ref{subsec:drift_detection_eval} depicts the performance of the drift detection models under a real life concept drift scenario. Section~\ref{subsec:human_feedback_eval} examines the improvements achieved through human feedback integration under the influence of concept drift.

\subsection{Baseline Agent Performance}\label{subsec:raw_performance}

The RL agent was initialized by injecting anomalies of three types --- sudden spikes, gradual drifts, and service degradation scenarios --- into the cluster services, generating a labeled dataset of over 800,000 data points. Of these, 300,000 were used for supervised training, 100,000 as a validation set to monitor for overfitting during training, and 400,000 as a held-out evaluation set. To increase precision of the reported values, the models were trained multiple times and the average was taken.

Table~\ref{tab:results} presents performance results comparing both agent variants against all individual detectors in the candidate pool. Individual detector performance was computed by applying each detector uniformly to all monitored time series and averaging results.

\begin{table}[t]
\centering
\caption{Anomaly detection performance comparison on held-out evaluation set.}
\label{tab:results}
\begin{tabular}{|l|c|c|c|}
\hline
\textbf{Method} & \textbf{F1} & \textbf{Recall} & \textbf{Precision} \\
\hline
F-DQN-Attn (proposed) & \textbf{0.6915} & \textbf{0.6755} & \textbf{0.7076} \\
F-DQN (MLP) & 0.4731 & 0.6439 & 0.3739 \\
\hline
MAD & 0.1071 & 0.5961 & 0.0588 \\
SRD & 0.0253 & 0.1310 & 0.0140 \\
SPOT & 0.0202 & 0.0104 & 0.4068 \\
OC-SVM & 0.0198 & 0.2181 & 0.0104 \\
RRCF & 0.0182 & 0.1301 & 0.0098 \\
LODA & 0.0085 & 0.0113 & 0.0068 \\
xStream & 0.0071 & 0.0037 & 0.1171 \\
\hline
\end{tabular}
\end{table}

The results highlight the limitation of applying any single fixed detector uniformly across heterogeneous services: individual detectors achieve F1 scores between 0.007 and 0.107, with no single method performing well across the full diversity of monitored time series. Detectors such as MAD achieve high recall at the cost of extremely low precision, indicating overfitting to specific anomaly patterns rather than generalization. SPOT exhibits the reverse pathology: high precision but near-zero recall, capturing only a narrow anomaly regime.

The MLP variant (F-DQN) achieves an F1 of 0.47, approximately fourfold the best individual detector, confirming that the learned selection policy substantially outperforms fixed assignment. The attention-augmented variant (F-DQN-Attn) further improves performance to F1 = 0.69, demonstrating that explicitly modeling inter-service dependencies --- the correlated behavioral patterns visible in the hub-and-spoke topology of Figure~\ref{fig:services} --- provides meaningful additional discriminative power.

\subsection{Drift Detection}\label{subsec:drift_detection_eval}

\begin{figure}[!tb]
    \centering
    \includegraphics[width=\linewidth]{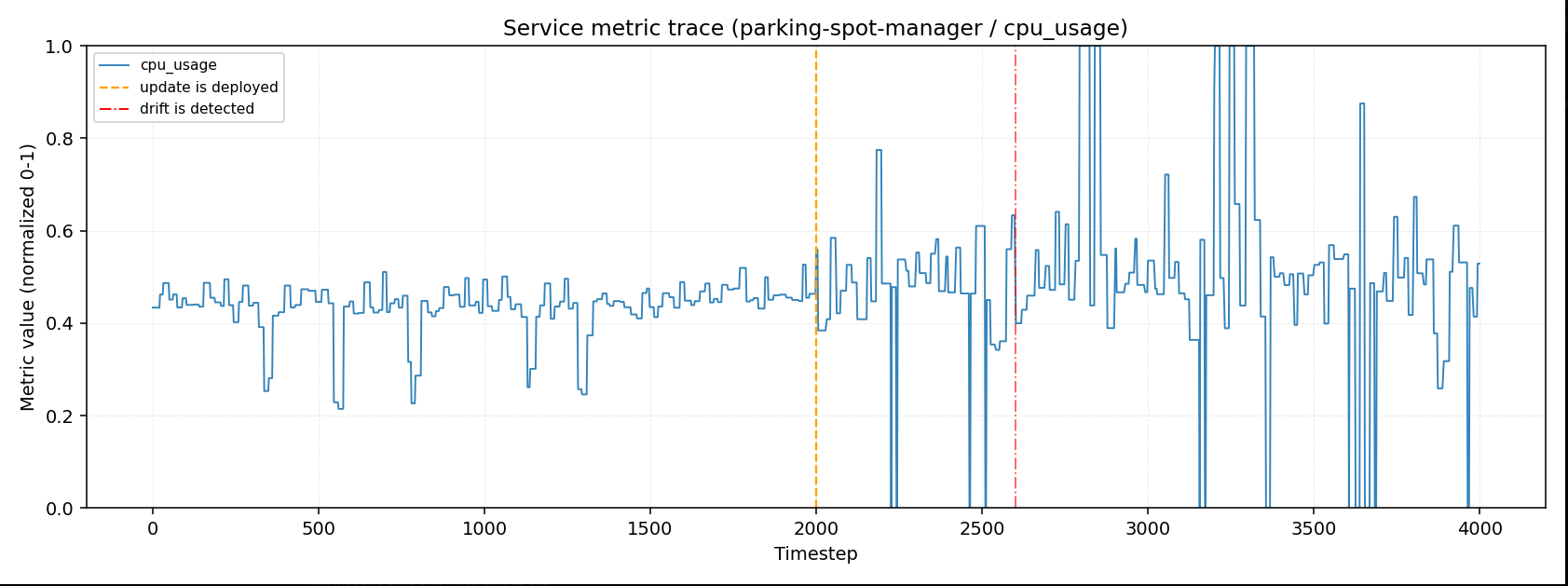}
    \caption{CPU utilization trace for the \texttt{parking-spot} service (normalized, 1-second resolution). The orange dashed line marks the deployment of a software update (step 2000). The red dash-dot line marks the point at which the drift detection ensemble reaches unanimous agreement (step 2600). Prior to the update, the signal is stationary with low variance (mean $\approx$ 0.45). Following the update, variance increases substantially, with frequent drops to zero and spikes approaching 1.0, clearly marking a distributional shift.}
    \label{fig:drift_res}
\end{figure}

To evaluate drift detection under realistic conditions, a software update was deployed to the AVP backend microservices. The update included operational stabilization patches, performance enhancements, and behavioral changes to the scheduling logic.

Figure~\ref{fig:drift_res} shows the CPU utilization trace for the \texttt{parking-spot} service. Prior to the update (steps 0--2000), the signal exhibits stationary behavior: low variance, mean approximately 0.45, bounded between 0.22 and 0.52. Following the update (step 2000 onward), the signal's character changes markedly: variance increases substantially, with frequent drops to zero (indicating periods of service inactivity) and peaks approaching 1.0 (indicating CPU saturation). This represents a simultaneous shift in mean behavior, marginal distribution shape, and multivariate dependency structure relative to the pre-update regime.

\begin{table}[b!]
\centering
\caption{Drift detector statistics before and after the software update, evaluated on the \texttt{parking-spot}/\texttt{cpu\_usage} time series. Values exceeding detection thresholds are marked in bold.}
\label{tab:results_drift}
\begin{tabular}{|l|c|c|}
\hline
\textbf{Drift detector} & \textbf{Pre-update} & \textbf{Post-update} \\
\hline
Page--Hinkley statistic & 4.4713 & \textbf{7.3299} \\
KS D-statistic & 0.1475 & \textbf{0.2234} \\
Mahalanobis outlier rate & 0.4110 & \textbf{0.6000} \\
\hline
\end{tabular}
\end{table}

Table~\ref{tab:results_drift} presents the statistics from all three drift detectors before and after the update. All three methods register values that exceed their respective detection thresholds in the post-update window, reaching unanimous agreement. The ensemble raises a drift alarm at step 2600, approximately 600 seconds (10 minutes) after update deployment. This detection latency is due to the ensemble's design: individual detectors signal change at different points, and the system waits for all three to concur before recommending retraining, trading a modest delay for high alarm precision.

\subsection{Expert Feedback and Retraining}\label{subsec:human_feedback_eval}

Following drift detection, the expert feedback interface was activated. Multiple error scenarios were injected into the post-update system, and the corresponding ground truth labels were used to simulate expert annotations. These labels were introduced through the pending transition buffer mechanism and used to trigger retraining with the 60/40 prioritized replay buffer strategy. Multiple retraining runs were conducted and the results were averaged.

\begin{figure}[tb]
    \centering
    \includegraphics[width=\linewidth]{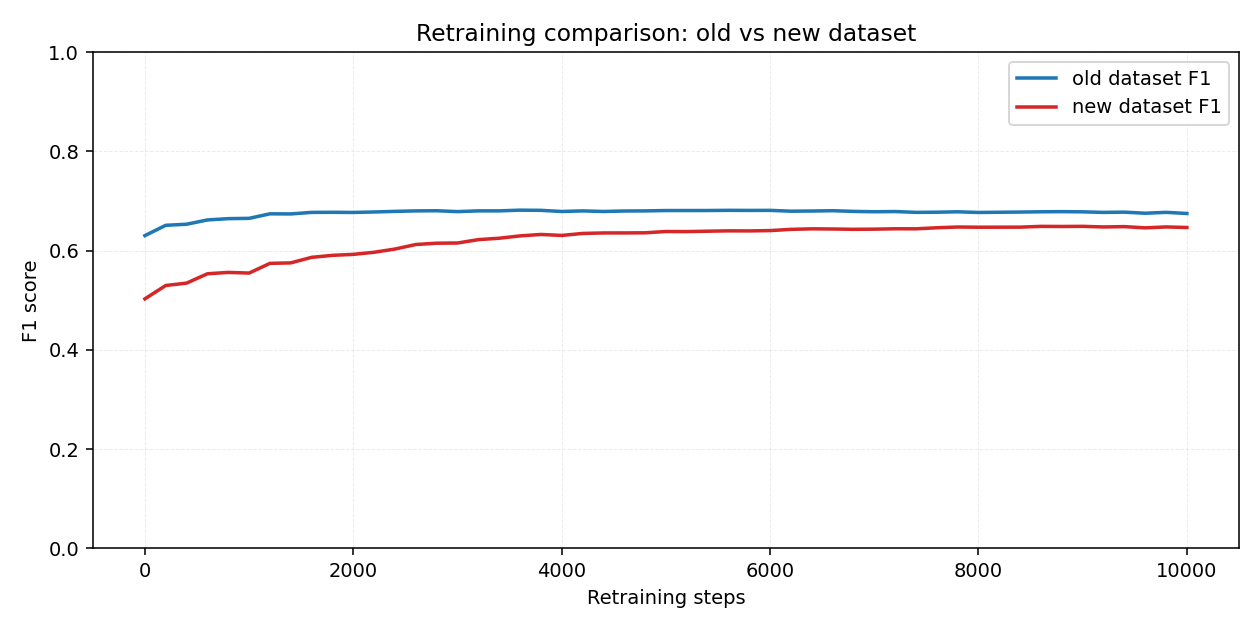}
    \caption{F1 score on old and new data distributions as a function of retraining steps. The 60/40 replay buffer sampling strategy (60\% new, 40\% old data) enables the model to adapt to the new distribution while preserving performance on the prior distribution throughout the retraining process.}
    \label{fig:retrain}
\end{figure}

Figure~\ref{fig:retrain} presents the F1 score on both the old and new data distributions as a function of retraining steps, and Table~\ref{tab:results_retrain} summarizes the key performance phases.

The retraining curve reveals a stable adaptation trajectory. On the new data distribution, F1 rises steadily from 0.52 at step 0 to approximately 0.65 by step 4,000--5,000, after which it stabilizes. On the old data distribution, F1 begins at 0.64 and rises slightly to 0.69 within the first 1,000 steps before stabilizing throughout the remaining 10,000 retraining steps. Critically, \emph{the old-distribution performance never degrades}: the 60/40 replay sampling strategy successfully maintains knowledge of the prior distribution while the model adapts to the new one.

This result demonstrates that the proposed retraining approach achieves stable adaptation without catastrophic forgetting. The convergence of both curves around step 4,000--5,000 suggests that the majority of adaptation occurs in the early retraining phase, with diminishing returns thereafter, providing a practical criterion for terminating the retraining loop.

\section{Conclusion}\label{sec:conclusion}
This paper presented self-adaptive anomaly detection framework for connected vehicle environments that integrates reinforcement-learning-based detector selection, statistical drift detection, and human-in-the-loop retraining into a single online supervisory loop. The framework was implemented within the SDVDiag platform and evaluated on an automated valet parking testbed across seven backend microservices.

The evaluation produced three principal findings. First, the attention-augmented factorized agent (F-DQN-Attn) reached an F1 score of 0.69 on the held-out evaluation set, substantially exceeding both the MLP variant (0.47) and any individual detector applied uniformly (at most 0.11), confirming that explicitly modeling inter-service dependencies provides meaningful discriminative power in the hub-and-spoke microservice topology. Second, the ensemble drift detector reached unanimous agreement approximately 10 minutes after a real software update, validating the conjunctive alarm rule as a precision-prioritizing trigger for retraining. Third, operator-triggered retraining with the 60/40 prioritized replay strategy recovered F1 to 0.65 on the post-update distribution while preserving 0.69 on the prior distribution, demonstrating online adaptation without catastrophic forgetting. Together, these results show that automated adaptation and operator oversight can be combined within a single coordinated loop rather than treated as competing design choices.

Several limitations bound the scope of these conclusions. For reproducibility, expert feedback was simulated by routing ground-truth labels through the feedback interface; this idealizes the operator and excludes annotation noise or expert disagreement that real deployments would introduce. The evaluation covers a single testbed, a single connected vehicle application, and a single observed concept drift event, leaving generalization across topologies, applications, and longer adaptation histories empirically open. Backend metrics were further limited to CPU and RAM at one-second resolution, and the injected anomaly catalog covered three pattern families, neither of which exhausts all failure modes of production CV systems.

Future work will address these limitations along three directions. First, the framework will be evaluated on additional testbeds, applications, and metric modalities, including in-vehicle and network signals, to characterize generalization beyond the present setup. Second, the simulated-expert protocol will be replaced with studies involving real operators, enabling investigation of annotation noise, disagreement, and the design of feedback interfaces that remain effective under realistic operator workloads. Third, statistical robustness and methodological depth will be strengthened through multi-seed evaluation, ablations on the drift-ensemble composition and replay ratio, and characterization of the attention layer's computational cost.

\begin{table}[t]
\centering
\caption{Anomaly detection performance across adaptation phases.}
\label{tab:results_retrain}
\begin{tabular}{p{0.28\linewidth} p{0.12\linewidth} p{0.50\linewidth}}
\hline
\textbf{Phase} & \textbf{F1} & \textbf{Remarks} \\
\hline
Pre-update (baseline) & 0.6915 & Performance on trained distribution (F-DQN-Attn). \\
\hline
Post-update, pre-retraining & 0.5239 & Concept drift degrades performance; new distribution causes false positives and false negatives. \\
\hline
Post-retraining, new dist. & 0.6500 & F1 on post-update data after retraining with human feedback. \\
\hline
Post-retraining, old dist. & 0.6900 & F1 on pre-update data after retraining; no catastrophic forgetting. \\
\hline
\end{tabular}
\end{table}

\section*{Acknowledgment}
The authors disclose that generative AI has been used to improve grammar and language of this publication. The authors have reviewed and edited all content as needed and take full responsibility for its scientific integrity and authenticity.

\bibliographystyle{IEEEtran}
\bibliography{bib}

\end{document}